\def\year{2021}\relax
\documentclass[letterpaper]{article} 
\usepackage{aaai21}  
\usepackage{times}  
\usepackage{helvet} 
\usepackage{courier}  
\usepackage[hyphens]{url}  
\usepackage{amsmath}
\usepackage{graphicx} 
\usepackage{natbib}  
\usepackage{caption} 
\usepackage{multirow}
\title{BERTKG-DDI: Towards Incorporating Entity-specific Knowledge Graph Information in Predicting Drug-Drug Interactions}

\author{
    Ishani Mondal \\
}
\affiliations{
    Microsoft Research Lab, \\
    Lavelle Road, Bengaluru, India \\
    t-imonda@microsoft.com/ishani340@gmail.com
}

\begin{document}
\maketitle

\begin{abstract}
 Off-the-shelf biomedical embeddings obtained from the recently released various pre-trained language models (such as BERT, XLNET) have demonstrated state-of-the-art results (in terms of accuracy) for the various natural language understanding tasks (NLU) in the biomedical domain. Relation Classification (RC) falls into one of the most critical tasks. In this paper, we explore how to incorporate domain knowledge of the biomedical entities (such as drug, disease, genes), obtained from Knowledge Graph (KG) Embeddings, for predicting Drug-Drug Interaction from textual corpus. We propose a new method, \emph{\textbf{BERTKG-DDI}}, to combine drug embeddings obtained from its interaction with other biomedical entities along with domain-specific BioBERT embedding-based RC architecture. Experiments conducted on the DDIExtraction 2013 corpus clearly indicate that this strategy improves other baselines architectures by 4.1\% macro F1-score.
\end{abstract}

\section{Introduction}
During the concurrent administration of multiple drugs to a patient, there seems to be a possibility in which an ailment might get cured or it can lead to serious side-effects. These type of interactions are known as Drug-Drug Interactions (DDIs). Predicting drug-drug interactions (DDI) is a difficult task as it requires to understand the underlying action principle of the interacting drugs. Numerous efforts by the researchers have been observed recently in terms of automatic extraction of DDIs from the textual corpus \cite{SAHU201815}, \cite{cnn}, \cite{Sun_2019}, \cite{li-ji-2019-syntax}, \cite{mondal-2020-bertchem} and predicting unknown DDI from KG \cite{8941958}. Automatic extraction of DDI from texts helps to maintain large-scale databases  and thereby facilitate the medical experts in their diagnosis.

In parallel to the progress of DDI extraction from the textual corpus, some efforts have been observed recently where the researchers came up with various strategies of augmenting chemical structure information of the drugs and textual description of the drugs \cite{ZHU2020103451} to improve Drug-Drug Interaction prediction performance from corpus and Knowledge Graphs. The DDI Prediction from the textual corpus has been framed by the earlier researchers as relation classification problem \cite{SAHU201815}, \cite{cnn}, \cite{Sun_2019}, \cite{li-ji-2019-syntax} using CNN or RNN-based neural networks. 

Recently, with the massive success of the pre-trained language models \cite{Devlin2019BERTPO}, \cite{Yang2019XLNetGA} in many NLP classifications, we formulate the problem of DDI classification as a relation classification task by leveraging both entities and contextual information. We propose a model that leverages both domain-specific contextual embeddings (Bio-BERT) \cite{biobert} from the target entities (drugs) and also its external information. In the recent years, representation learning has played a pivotal role in solving various machine learning tasks. 

In this work, we explore the direction of augmenting graph embeddings to predict relation between two drugs from the textual corpus. We have made use of an in-house Knowledge Graph (\emph{Bio-KG}) after curating the interactions among drugs, diseases, genes from multiple ontologies. 
In order to understand the complex underlying mechanism of interactions among the biomedical entities, we employ translation-based and semantics preserving heterogeneous graph embeddings on \emph{Bio-KG} and augment the entities representation jointly to train the relation classification model. Experiments conducted on the DDIExtraction 2013 corpus \cite{HERREROZAZO2013914} reveals that this method outperforms the existing baseline models and is in line with the new direction of research of fusing various information to DDI prediction. In a nutshell, the major contributions of this work are summarized as follows:
\begin{enumerate}
    \item  We propose a novel method that jointly leverages textual and external Knowledge information to classify relation type between the drug pairs mentioned in the text showing the efficacy of external entity specific information.
    \item Our method achieves new state-of-the-art performance on DDI Extraction 2013 corpus.
\end{enumerate}

\section{Problem Statement}
Given an input instance or sentence $s$ with two target drug entities $d_1$ and $d_2$, the task is to classify the type of relation ($y$) the drugs hold between them, $y$ $\in$ ($y_1$ , ...., $y_N$). Here $N$ denotes the number of relation types.

\section{Methodology}
\subsection{Text-based Relation Classification}

Our model for extracting DDIs from texts is based on the pre-trained BERT-based relation classification model by \cite{rbert}. Given a sentence $s$ with drugs $d_{1}$ and $d_{2}$, let the final hidden state output from BERT module is $H$. Let the vectors $H_{i}$ to $H_{j}$ are the final hidden state vectors from BERT for entity $d_{1}$, and $H_{k}$ to $H_{m}$ are the final hidden state vectors from BERT for entity $d_{2}$. 
An average operation is applied to obtain the vector representation for each of the drug entities. An activation operation tanh is applied followed by a fully connected layer to each of the two vectors, and the output for $d_1$ and $d_2$ are $H_{1}^{'}$ and $H_{2}^{'}$ respectively. 

\begin{equation}
    H_{1}^{'} = W [tanh(\frac{1}{(j-i+1)}\sum_{t=i}^{j} H_t] + b
\end{equation}
\begin{equation}
    H_{2}^{'} = W [tanh(\frac{1}{(m-k+1)}\sum_{t=k}^{m} H_t] + b
\end{equation}

\noindent
The weights ($W$) and bias ($b$) parameters are shared. For the final hidden state vector of the first token (‘[CLS]’), we also add an activation operation and a fully connected layer, which is formally expressed as:

\begin{equation}
    H_{0}^{'} = W_0 (tanh(H_0)) + b_0
\end{equation}
Matrices $W_0$, $W_1$, $W_2$ have the same dimensions, i.e. $W_0$ $\in$ $R^{d * d}$ ,$W_1$ $\in$  $R^{d * d}$, $W_2$ $\in$  $R^{d * d}$, where $d$ is the hidden state size from BERT. 
\noindent
We concatenate $H_{0}^{'}$, $H_{1}^{'}$ and $H_{2}^{'}$ and then add a fully connected layer and a softmax layer, which is expressed as :

\begin{equation}
    h^{''} = W_3 [concat(H_{0}^{'}, H_{1}^{'}, H_{2}^{'})] + b_3
\end{equation}
\begin{equation}
    y_t^{'} = softmax(h^{''})
\end{equation}
 
\noindent
$W_3$ $\in$ $R ^ {N*3d}$, and $y_t^{'}$ is the softmax probability output over $N$. In Equations (1), (2), (3), (4) the bias vectors are $b_0$, $b_1$, $b_2$, $b_3$. We use cross entropy as the loss function. We denote this text-based architecture as \emph{BERT-Text-DDI}.

\subsection{Entity Representation from KG}
To infuse external information of the entities in relation classification task, we obtain the representation of two \emph{Drug} entities mentioned in each input instance of the relation classification task. We use an in-house heterogeneous biomedical Knowledge Graph (\emph{Bio-KG}) consisting of the interactions of target-target, drug-drug, drug-disease, drug-target, disease-disease, disease-target interactions from a large number of ontologies such as  : DrugBank\footnote{https://go.drugbank.com/}, BioSNAP\footnote{http://snap.stanford.edu/biodata/}, UniProt\footnote{https://www.uniprot.org/} \cite{10.1093/nar/gkw1099}. The overall statistics of \emph{Bio-KG} has been enumerated in table 1. The real-world information/facts observed in the \emph{Bio-KG} are stored as a collection of triples in the form {($h$, $r$, t)}. Each triple is composed of a head entity $h$ $\in$ $E$, a tail entity $t$ $\in$ $E$, and a relation $r$ $\in$ $R$ between them, e.g., (\emph{paracetamol}, \emph{treats}, \emph{fever}). The fact that \emph{paracetamol} is effective in curing \emph{fever} is being stored in \emph{Bio-KG}. In this case, $E$ denotes set of entities, and $R$ denotes the set of relations. There are three different types of $E$ in \emph{Bio-KG} such as \emph{drugs, diseases, targets} and five different types of $R$ such as \emph{target-target, drug-disease, drug-target, disease-disease, disease-target interactions}.

\begin{table}[!t]
\centering
 \begin{tabular}{|c c | c c|} 
 \hline
 Node Types & Count & Edge Types & Count \\ [0.5ex] 
 \hline
 Drug  & 6512 & Drug-Target & 15245 \\ 
 Target & 30098 & Target-Target & 77108 \\
 
  Disease & 23458 & Drug-Disease & 84745 \\
  
   & & Disease-Disease & 35382 \\
   
   &  & Disease-Target & 31161 \\
   
 \hline
  Total Nodes & 60068 & Total Edges & 243641\\
  [1ex] 
 \hline
 
 \end{tabular}
 \label{stats}
 \caption {Statistics of Bio-KG.}
\end{table}

The aim of a Knowledge Graph embedding is to embed the entities and relations into a low-dimensional continuous vector space, so as to simplify the computations on the KG. They mostly use facts in the KG to perform the embedding task, enforcing embedding to be compatible with the facts. They provide a generalizable context about the overall Knowledge Graph (KG) that can be used to infer the relations. In this work, we employ some off-the-shelf KG embeddings to encode the representation of each of the drugs (in terms of their relationship with other entities). The knowledge graph embeddings are computed so that they satisfy certain properties; i.e., they follow a given KGE model. These KGE models define different score functions that measure the distance of two entities relative to its relation type in the low-dimensional embedding space. These score functions are used to train the KGE models so that the entities connected by relations are close to each other while the entities that are not connected are far away.
Some of the KGEs used in our experiments as explained below:

\begin{itemize}
    \item \textbf{TransE} \cite{transe}: Given a fact ($h$, $r$, $t$), the relation in TransE is interpreted as a translation vector $r$ so that the embedded entities $h$ and $t$ can be connected by $r$, i.e., $h$ + $r$ $\approx$ $t$ when ($h$, $r$, $t$) holds. The scoring function is defined as (negative) distance between $h + r$ and $t$, i.e.,
    \begin{equation}
        f_r(h, t) = \parallel h + r - t \parallel
    \end{equation}

\item \textbf{TransR}  \cite{transr}: Given
a fact ($h$, $r$, $t$), TransR first projects the entity representations $h$ and $t$ into the space specific to relation $r$, Here $M_r$ is a projection matrix from the entity space
to the relation space of $r$, the scoring function is:
\begin{equation}
    h_t = M_r h  , t_t = M_rt
\end{equation}
\item \textbf{RESCAL} \cite{rescal}: Each relation in RESCAL is represented as a matrix which models pairwise interactions between latent factors. The score of a fact ($h$, $r$, $t$) is defined by a bi-linear function where $h$, $t$ are vector representations of the entities, and $M_r$ is a matrix associated with the relation.
This score captures pairwise interactions between all
components of $h$ and $t$:
\begin{equation}
    f_r(h, t) = h^{T}M_rt = \sum_{i=0}^{d-1} \sum_{j=0}^{d-1} [M_r]_{ij} * [h]_i * [t]_j
\end{equation}

\item \textbf{DistMult} \cite{distmult}: DistMult simplifies RESCAL by restricting $M_{r}$ to diagonal matrices. For each relation $r$, it introduces a
vector embedding $r$ and requires $M_{r}$ = $diag(r)$. The
scoring function is defined as:
\begin{equation}
    f_r(h, t) = h^{T}diag(r)t = \sum_{i=0}^{d-1} r_i * [h]_i * [t]_j
\end{equation}
This score captures pairwise interactions between only the
components of $h$ and $t$ along the same dimension, and reduces the number of parameters to $O(d)$ per relation.

\end{itemize}

\noindent
From \emph{Bio-KG}, we train these KG Embeddings and obtain the representation of all the nodes. In our case, we are only interested in obtaining the representation of \emph{drug} nodes. We denote the KG representation of drug $d$ as ${kge}$.

\subsection{BERTKG-DDI}
From the input instance $s$ with two tagged target drug entities $d_1$ and $d_2$, we obtain the KG embedding representation of two drugs ${kge}_1$ and ${kge}_2$ respectively using \emph{Bio-KG}. We concatenate these two embeddings ${kge}_1$ and ${kge}_2$ and pass those through a fully connected layer as represented below:

\begin{equation}
    kge = W [concat({kge}_1, {kge}_2)] + b
\end{equation}

\noindent
$W$ and $b$ are the parameters of the fully-connected layer of the KG representation of ${kge}_1$ and ${kge}_2$. The final layer of \emph{BERTKG-DDI model} contains concatenation of all the previous text-based outputs and drug representation from KG as expressed below:

\begin{equation}
    o^{'} = W_3 [concat(H_{0}^{'}, H_{1}^{'}, H_{2}^{'}, kge)] + b_3
\end{equation}
\begin{equation}
    y_t^{'} = softmax(o^{'})
\end{equation}

\noindent
Finally the training optimization is achieved using the cross-entropy loss.
    
\section{Experimental Setup}
\subsection{Dataset and Pre-processing}
We have followed the task setting of Task 9.2 in the
DDIExtraction 2013 shared task \cite{HERREROZAZO2013914} for evaluation. It consists of MEDLINE documents annotated with the drug mentions and five types of interactions: \emph{Mechanism}, \emph{Effect}, \emph{Advice}, \emph{Interaction} and \emph{Other}. The task is a multi-class classification to classify each of the drug pairs in the sentences into one of the types and we evaluate using three standard evaluation metrics such as: Precision (P), Recall (R) and F1-score (F1). 

During pre-processing, we obtain the DRUG mentions in the corpus and map those into unique DrugBank \footnote{https://go.drugbank.com/} identifiers. This is a step for converting the drug mentions into their respective DrugBank ID, a step of entity linking \cite{mondal-etal-2019-medical}, \cite{dnorm}. This mention normalization has been performed based on the longest overlap of drug mentions in DrugBank and map the drugs to different Knowledge sources used to construct \emph{Bio-KG}.

\subsection{Training Details}
For the purpose of experiments, we use the initialization of various pre-trained contextual embeddings. For instance, we use the embeddings such as \emph{\textbf{bert-base-cased}} \footnote{https://huggingface.co/bert-base-cased}, \emph{\textbf{scibert-scivocab-uncased}} \cite{Beltagy2019SciBERTAP} \footnote{https://github.com/allenai/scibert} and domain-specific \emph{\textbf{biobert v1.0 pubmed pmc}} and \emph{\textbf{biobert v1.0 pubmed}}\footnote{https://github.com/dmis-lab/biobert} as the initialization of the transformer encoder in \textbf{BERTKG-DDI}. We uniformly keep the maximum sequence length as 300 for all the embedding ablations and trained for 5 epochs. For the KG embeddings, we use word embeddings dimensions to be 200. Stochastic Gradient Descent (SGD) was used for optimization with an initial learning rate of 0.0001 and the model is trained for 300 epochs. After training the embeddings, we obtain the final representation of each drug. For the drugs mentioned in the input instance, we make use of the obtained embeddings as shown in the equation 11. We initialize the non-normalized drugs using pre-trained word2vec (of dimension 200 same as the KG embedding) trained on PubMED \footnote{http://evexdb.org/pmresources/ngrams/PubMed/}.

\begin{table}[!t]
\small
\begin{center}
\begin{tabular}{|c|c|}
\hline
\textbf{Embeddings on BERT-Text-DDI} & \textbf{Test set Macro F1}\\ 
\hline
\emph{bert-base-cased} & 0.806 \\
\emph{scibert-scivocab-uncased} & 0.812\\
\emph{biobert v1.0 pubmed pmc} &  0.818 \\
\emph{biobert v1.1 pubmed} & 0.822 \\
\hline
\end{tabular}
\end{center}
\caption{Ablation of the contextual embeddings.}
\label{berttext-ddi}
\end{table}

\begin{table}[!t]
\small
\begin{center}
\begin{tabular}{|c|c|}
\hline
\textbf{KG Embeddings on BERTKG-DDI} & \textbf{Test set Macro F1}\\ 
\hline
\emph{BERTKG-DDI w/ TransE} & 0.826 \\
\emph{BERTKG-DDI w/ TransR} & 0.829 \\
\emph{BERTKG-DDI w/ RESCAL} &  0.834 \\
\emph{BERTKG-DDI w/ DistMult} & 0.840 \\
\hline
\end{tabular}
\end{center}
\caption{Ablation of the influence of KG embeddings.}
\label{bertkg-ddi}
\end{table}

\begin{table}[!t]
\small
\begin{center}
\begin{tabular}{|c|c|c|}
\hline
\textbf{Models} & \textbf{Contextual Embeddings} & \textbf{Macro F1}\\ 
\hline
\emph{BERT-Text-DDI} & \emph{biobert v1.0 pubmed pmc} &  0.818 \\
\emph{BERTKG-DDI} & \emph{biobert v1.0 pubmed pmc} &  0.831 \\
\emph{BERT-Text-DDI} & \emph{biobert v1.1 pubmed} & 0.822 \\
\emph{BERTKG-DDI} & \emph{biobert v1.1 pubmed} & 0.840 \\
\hline
\end{tabular}
\end{center}
\caption{Probing deeper into influence of KG embeddings into BERT-based models for DDI Relation Classification.}
\label{berttextandkg-ddi}
\end{table}

\begin{table*}[!t]
\small
\begin{center}
\begin{tabular}{|c|c|c|c|c|c|c|c|c|c|}
\hline
\multirow{2}{*}{\textbf{Methods}}& \textbf{Advice}&\textbf{Effect}&\textbf{Mechanism}&\textbf{Interaction}&\textbf{Total}\\ 
& \textbf{F1 Score}  & \textbf{F1 Score} & \textbf{F1 Score} & \textbf{F1 Score} & \textbf{F1 Score}\\ \hline

\cite{zhang}  & 0.80 & 0.71 & 0.74 & 0.54 & 0.72 \\
\cite{zheng}  & 0.85 & 0.76 &  0.77  & 0.57 & 0.77 \\
\cite{asada-etal-2018-enhancing}  & 0.81 & 0.71 & 0.73 & 0.45 & 0.72 \\
\cite{Sun_2019}  & 0.80 & 0.73 & 0.78  & 0.58 & 0.75 \\
\cite{ZHU2020103451} & 0.86 & 0.80 & 0.84 & 0.56 & 0.80\\
\hline
Our method (BERTKG-DDI) & \textbf{0.88} & \textbf{0.81} & \textbf{0.87} &  \textbf{0.59} & \textbf{0.84} \\
\hline
\end{tabular}
\end{center}
\caption{Comparison of F1 scores across relation types using baselines on the test set of the DDI Benchmark Corpus, containing five types of relations between drug pairs (\emph{Advice, Mechanism, Effect, Interaction, Others}.}
\label{final}
\end{table*}

\section{Results and Discussion}
In this section, we provide a detailed analysis of the various results and findings that we have observed during experiments. We show empirical results based on \emph{BERTKG-DDI} for both text and KG information.\\

\noindent
\textbf{Ablation of Embeddings on BERT-Text-DDI}:
During ablation analysis, we observe that the incorporation of domain-specific information in \emph{biobert v.1 pubmed} boosts up the predictive performance in terms of macro-F1 score (across all relation types) by 2.3\% compared to \emph{bert-base-cased}. Moreover, the \emph{scibert-vocab-cased} embedddings due to the scientific details obtained during fine-tuning achieves reasonable boost in performance. \emph{biobert v.1 pubmed based BERT-Text-DDI} is the best-performing text-based relation classification model. The results are enumerated in Table \ref{berttext-ddi}.\\

\noindent
\textbf{Ablation analysis of KG Embeddings on BERTKG-DDI:}
We compare the different KG embeddings for drugs obtained from \emph{Bio-KG} after augmenting with the \emph{BERT-Text-DDI} model in Table \ref{bertkg-ddi}. The semantic-matching models such as \emph{RESCAL} and \emph{DistMult} measure plausibility of facts by matching the latent semantics of both relations and entities in their vector space. In our experiments, they seem to outperform the translation-based KGE such as \emph{TransE} and \emph{TransR} by an average of 1\% macro F1-score.\\

\noindent
\textbf{Advantage of KG information on BERTKG-DDI:}
During empirical analysis of the \emph{BERTKG-DDI} model, we observe how much performance gain can be achieved by augmenting KG embeddings. From the results enumerated in terms of macro F1-score on all the relation types in Table \ref{berttextandkg-ddi}, we observe that the best-performing \emph{BERT-Text-DDI} model achieves a performance boost of 1.8\% after augmenting KG information in \emph{BERTKG-DDI}. \\ 

\noindent
\textbf{Comparison with the existing baselines: }
We compare our best-performing model with some of the best-performing existing baselines. \cite{asada-etal-2018-enhancing} proposed a novel neural method to extract drug-drug interactions (DDIs) from texts using external drug molecular structure information. They encode textual drug pairs with convolutional neural networks and their molecular pairs with graph convolutional networks (GCNs), and then concatenate the outputs of these two networks. \cite{zheng} proposed an effective model that classifies DDIs from the literature by combining an attention mechanism and a recurrent neural network with long short-term memory (LSTM) units. \cite{zhang} has presented a hierarchical recurrent neural networks (RNNs)-based method to integrate the SDP and sentence sequence for DDI extraction task. 
\cite{Sun_2019} has proposed a novel recurrent hybrid convolutional neural network (RHCNN) for DDI extraction from biomedical literature. In the embedding layer, the texts mentioning two entities are represented as a sequence of semantic embeddings and position embeddings. In particular, the complete semantic embedding is obtained by the information fusion between a word embedding and its contextual information which is learnt by recurrent structure.  Recently, \cite{ZHU2020103451} proposed multiple entity-aware attentions with various entity information to strengthen the representations of drug entities in sentences. They integrate drug descriptions from Wikipedia and DrugBank to our model to enhance the semantic information of drug entities. Also, they modified the output of the BioBERT model and the results show that it is better than using the BioBERT model directly. On the contrary, our method achieves the state-of-the-art performance based on the results on the DDI Extraction 2013 corpus (in terms of F1-scores of all the relation types) as shown in Table \ref{final}.

\section{Conclusion}
In this paper, we propose an approach, \emph{BERTKG-DDI}, for DDI relation classification based on pre-trained language models and Knowledge Graph Embedding of the drug entities. Experiments conducted on a benchmark DDI dataset proves the effectiveness of our proposed method. Possible directions of further research might be to explore other external drug representation such as chemical structure, textual description in predicting DDI from textual corpus.
\bibliography{aaai}
\end{document}